\documentclass[runningheads]{llncs}
\usepackage[T1]{fontenc}
\usepackage[british]{babel}
\usepackage{graphicx}
\usepackage{booktabs}
\usepackage{mathtools}
\usepackage{xspace}
\usepackage{multirow}
\usepackage{caption}
\usepackage{subcaption}
\usepackage{capt-of}
\usepackage{amssymb}
\usepackage{dsfont}

\newcommand{\ie}{\emph{i.e., \xspace}}

\newcommand{\wrt}{\emph{{w.r.t.} \xspace }}
\newcommand{\eg}{\emph{e.g.,\xspace}\xspace}

\DeclareMathOperator*{\argmax}{arg\,max}

\newcommand{\particul}{PARTICUL\xspace}

\usepackage{wasysym}
\usepackage{color, colortbl}

\usepackage{hyperref}
\usepackage{color}

\newcommand\citep{\cite}
\begin{document}
\title{Contextualised Out-of-Distribution Detection using Pattern Identification}
\titlerunning{Contextualised OoD Detection using Pattern Identification}
%
\author{
	Romain Xu-Darme\inst{1,3} \and
	Julien Girard-Satabin\inst{1} \and
	Darryl Hond\inst{2} \and
	Gabriele Incorvaia\inst{2} \and
	Zakaria Chihani\inst{1}
}
\authorrunning{R.Xu-Darme et al.}
\institute{
	Université Paris-Saclay, CEA, List, F-91120, Palaiseau, France \\
	\email{<first>.<last>@cea.fr}
	\and
	Thales UK, Research, Technology and Innovation, Reading, UK \\
	\email{<first>.<last>@uk.thalesgroup.com}
	\and
	Univ. Grenoble Alpes, CNRS, Grenoble INP, LIG, F-38000 Grenoble, France
}

\maketitle              
\begin{abstract}
In this work, we propose CODE, an extension of existing work from the field of explainable AI
that identifies class-specific recurring patterns
to build a robust Out-of-Distribution (OoD) detection method 
for visual classifiers.
CODE does not 
require any classifier retraining and is OoD-agnostic, \ie tuned directly to the training dataset.
Crucially, pattern identification allows us to provide images from
the In-Distribution (ID) dataset as reference data to provide additional context to  
the confidence scores.
 
In addition, we introduce a new benchmark based on perturbations of the ID dataset that provides
a known and quantifiable measure of the discrepancy between the ID and OoD datasets 
serving as a reference value for the comparison between OoD detection methods.

\keywords{Out-of-distribution detection  \and Explainable AI \and Pattern identification}
\end{abstract}
\section{Introduction}
A fundamental aspect of software safety is arguably the
modelling of its expected operational
domain through a formal or semi-formal specification, giving clear
boundaries on when it is sensible to deploy the program, and
when it is not.
It is however difficult to define such boundaries
for machine learning programs, especially for visual classifiers based on
artificial neural networks (ANN) that 
process high-dimensional data (images, videos) and are the result of a complex optimisation procedure.
In this context, Out-of-Distribution (OoD)
detection - which aims to detect whether an input of an ANN is
In-Distribution (ID) or outside of it - serves several purposes: 1) it helps
characterise the extent to which the ANN can operate outside a bounded dataset;
2) it constitutes a surrogate measure of the
generalisation abilities of the ANN; 3) it can help assess when
an input is too far away from the operational domain, which prevents misuses of
the program and increases its safety.
However, one crucial aspect missing from current OoD detection methods is the ability to
provide some form of explanation of their decision. Indeed, most approaches are based
on a statistical model of the system behaviour, built upon an abstract
representation of the input data, sometimes turning OoD detection into an opaque decision that may appear arbitrary to the end-user.
While it would be possible to generate a visual representation of the abstract space using tSNE and to highlight ID data clusters for justifying the OoD-ness of a given sample, tSNE is extremely dependent on the choice of hyper-parameters, sometimes generating misleading visualisations~\cite{wattenberg2016how}.
In this regard, methods from the field of Explainable AI (XAI), which are typically used
to provide some insight about the decision-making process of the model, can be adapted to
build models for OoD detection that provide some context information to justify their decision.
In the particular task of image classification, XAI methods can help extract
visual cues that are class-specific (\eg a bird has wings), and whose presence or absence can
help characterise the similarity of the input image to the target distribution (\eg an
object classified as a bird that shows neither wings nor tail nor beak is probably an OoD input). Therefore, in this work we make the following contributions:
\begin{enumerate}
	\item We introduce a new benchmark based on perturbations of the ID dataset which provides
		a known and quantifiable evaluation of the discrepancy between the ID and OoD datasets
		that serves as a reference value for the comparison between various OoD detection methods (Sec.~\ref{sec:perturbation}).
	\item We propose CODE, an OoD agnostic detection measure that does not require any fine-tuning of the original classifier. Pattern identification allows us to provide images from the ID dataset as reference points to justify the decision (Sec.~\ref{sec:code}). Finally, we demonstrate the capabilities of this approach in a broad comparison with existing methods (Sec.~\ref{sec:experiments}).
\end{enumerate}

\section{Related Work}\label{sec:sota}
\paragraph{Out-of-distribution detection.}
In this work, we focus on methods that can apply to pre-trained classifiers. Therefore, we exclude methods
which integrate the learning of the confidence measure within the training objective of the model,
or specific architectures from the field of Bayesian Deep-Learning that aim at capturing uncertainty by design. 
Moreover, we exclude \emph{OoD-specific} methods 
that use a validation set composed of OoD samples
for the calibration of hyper-parameters, and focus on \emph{OoD-agnostic} methods that require
only ID samples. \\
In this context, the maximum softmax probability (MSP) obtained after normalisation of the classifier logits constitutes a good baseline for OoD detection~\citep{hendrycks2017baseline}. %

More recently, ODIN~\citep{liang2018enhancing} measures the local stability of the classifier using gradient-based perturbations, while MDS~\citep{lee2018simple} uses the Mahalanobis distance to
class-specific points in the feature space.
\cite{liu2020energy} proposes a framework based on energy scores,
which is extended in the DICE method~\citep{sun2021dice} by first performing a class-specific directed sparsification of the last layer of the classifier.
ReAct~\citep{sun2021react} also modifies the original classifier by rectifying the activation values of the penultimate layer of the model.
\cite{hendrycks2022scaling} proposes two related methods: MaxLogit - based on the maximum logit value - and KL-Matching which measures the KL divergence between the output of the model and the class-conditional mean softmax values. 
The Fractional Neuron Region Distance~\citep{hond2021integrated} (FNRD) computes the range of activations for each neuron over the training set in order to empirically characterise the
statistical properties of these activations, then provides a score describing how many neuron outputs are outside the corresponding range boundaries for a given input.
Similarly, for each layer in the model, \cite{sastry2020detecting} computes the range of pairwise feature correlation between channels across the training set.
ViM~\citep{wang2022vim} adds a dedicated logit for measuring the OoD-ness of an input by using the residual of the feature against the principal space. 
KNN~\citep{sun2022out} uses the distance of an input to the $k$-th nearest neighbour.
Finally, GradNorm \citep{huang2021importance} measures the gradients of the
cross-entropy loss \wrt the last layer of the model.
\paragraph{Evaluation of OoD detection.}
All methods presented above are usually evaluated on different settings (\eg different ID/OoD datasets), sometimes using only low resolution images (\eg MNIST~\citep{deng2012mnist}), which only gives a partial picture of their robustness.
Therefore, recent works such as \citep{hendrycks2022scaling,sun2021dice} - that
evaluate OoD methods on datasets with higher resolution (\eg ImageNet~\citep{deng2009imagenet}) - or Open-OoD~\citep{yang2022openood} - which aims at standardising the evaluation of OoD detection, anomaly detection and open-set recognition into a unified
benchmark - are invaluable.
However, when evaluating the ability of a method to discriminate ID/OoD datasets, it is often difficult to properly quantify the
margin between these two datasets, \textit{independently} from the method under test, and
to establish a ``ground truth'' reference scale for this margin.
Although \cite{yang2022openood} distinguishes
\textit{“near-OoD datasets [that] only have semantic shift compared with ID datasets”}
from \textit{“far-OoD [that] further contains obvious covariate (domain) shift}~”, this taxonomy 
lacks a proper way to determine, given two OoD datasets, which is ``further'' from the
ID dataset. Additionally, \cite{mukhoti2022raising} generates
“\textit{shifted sets}” that are \textit{“perceptually dissimilar but semantically similar to the
training distribution”}, using a GAN model 
for measuring the \textit{perceptual} similarity, and a deep ensemble model for evaluating the
\textit{semantic} similarity between two images. However, this approach requires the training of
multiple models in addition to the classifier.
Thus, in this paper we propose a new benchmark based on gradated perturbations
of the ID dataset. This benchmark measures the correlation between the OoD detection score
returned by a given method when applied to a perturbed dataset (OoD), and the intensity of the corresponding perturbation.
\paragraph{Part detection.}
Many object recognition methods have focused on part detection, in supervised (using
annotations~\citep{zhao2019recognizing}), weakly-supervised (using class
labels~\citep{li2020attribute}) or unsupervised~
\citep{han2022pcnn,xudarme2022particul,zheng2017learning} settings, primarily with the goal
of improving accuracy on hard classification tasks. %
To our knowledge, the \particul algorithm~\citep{xudarme2022particul} is
the only method that includes a confidence measure associated with the detected parts
(used by the authors to infer the visibility of a given part). \particul aims
to identify \textit{recurring patterns} in the latent representation of a set of images
processed through a pre-trained CNN, in an unsupervised manner. It is, however, restricted
to homogeneous datasets where all images belong to the same macro-category.
For more heterogeneous datasets, it becomes difficult to find recurring patterns
that are present across the entire training set.


\section{Beyond cross-dataset evaluation: measuring consistency against perturbations}\label{sec:perturbation}
In this section, we present our benchmark for evaluating the consistency of OoD detection methods using perturbations of the ID dataset.

Let $f:\mathcal{X}\rightarrow \mathbb{R}^N$ be a classifier trained on a dataset
${X_{train}\sim\mathcal{P}_{id}}$, where $\mathcal{P}_{id}$ is a distribution over 
$\mathcal{X}\times \mathbb{R}^N$ and $N$ is the number of categories 
learned by the classifier. We denote $\mathcal{D}_{id}$ the marginal distribution of 
$\mathcal{P}_{id}$ over $\mathcal{X}$. For any image $x\in \mathcal{X}$, $f$ outputs a vector 
of logits $f(x)\in \mathbb{R}^N$. 
The index of the highest value in $f(x)$ corresponds to the most 
probable category (or \textit{class}) of $x$ - relative to all other categories. 

Without loss of generality, the goal of an OoD detection method is to build 
a class-conditional\footnote{Class-agnostic methods simply ignore the image label/prediction.}
confidence function $C: \mathcal{X}\times \mathbb{R}^N \rightarrow \mathbb{R}$ assigning 
a score to each pair $(x, y)$, where $y$ can be either the ground truth label of $x$
when known, or the prediction $f(x)$ otherwise. 
This function constitutes the basis of OoD detection,
under the assumption that images belonging to 
$\mathcal{D}_{id}$ should have a higher confidence score than images outside $\mathcal{D}_{id}$. 

A complete evaluation of an OoD detection method would
require the application of the
confidence function $C$
on samples representative of the ID and OoD distributions. However, it is not 
possible to obtain a dataset representative of all possible OoD inputs. Instead, \textbf{cross-dataset OoD evaluation}
consists in drawing a test dataset $X_{test}\sim \mathcal{D}_{id}$ (with $X_{test} \neq X_{train}$), 
choosing a different dataset $D_{ood} \not\sim \mathcal{D}_{id}$,
then measuring
the \textit{separability} of $C(X_{test})$ and $C(D_{ood})$, where $C(X)$ denotes the 
distribution of scores computed over dataset $X$ using $C$. 
Three metrics are usually used:
Area Under the ROC curve (AUROC);
Area Under the Precision-Recall curve (AUPR),
False Positive Rate when the true positive rate is 95\% (FPR95).

In this work, in addition to cross-dataset evaluation, 
we propose to \emph{generate} an OoD distribution $\mathcal{D}_{ood}$ by applying a perturbation 
to all images from $\mathcal{D}_{id}$. Although image perturbation is a standard method
for evaluating the robustness of classifiers~\cite{hendrycks2018benchmarking}, our intent differs:
rather than trying to capture the point of failure of a classifier,
we monitor how the various confidence scores evolve when applying a perturbation of
increasing intensity to the ID dataset.
In practice, we use four transformations: Gaussian noise, Gaussian blur, brightness changes and rotations.
More generally, a perturbation $P_\alpha$ is a function
that applies a transformation of magnitude $\alpha$ to an image $x\in \mathcal{X}$
(\eg a rotation with angle $\alpha$).
When applying $P_\alpha$ over $\mathcal{D}_{id}$, we define the expected confidence as
\begin{equation}\label{eq:avg_confidence_perturbation}
	E(P_\alpha,C)=\mathbb{E}_{x\sim \mathcal{D}_{id}} \big[C\big(P_\alpha(x),f(P_\alpha(x))\big)\big]
\end{equation}
which is evaluated over the test set $X_{test}$. 
Although it would again be possible to measure the separability of ID and OoD confidence distributions, 
perturbations of small magnitude would result in almost identical distributions.
Instead, we evaluate the correlation between the magnitude of the perturbation and the average confidence 
value of the perturbed dataset as the Spearman Rank Correlation Coefficient (SRCC) $r_s$ between $\alpha$ 
and $E(P_\alpha,C)$, using multiple magnitude values $(\alpha_0, \ldots, \alpha_n)$.
$r_s=1$ (resp. $-1$) indicates that the average confidence measure
increases (resp. decreases) \textit{monotonically} with the value of
$\alpha$, \ie that the measure is
correlated with the magnitude of the perturbation. 
The key advantage of the SRCC resides in the ability to compare the general behaviour of various OoD detection methods that usually have different calibrations (\ie different range of values).
Assuming that the discrepancy between $\mathcal{D}_{id}$ and 
$P_\alpha(\mathcal{D}_{id})$ is correlated to the
magnitude of the perturbation $\alpha$ (ground truth), this benchmark measures the \textit{consistency} of the OoD methods under test.

\section{Contextualised OoD Detection using Pattern Identification}\label{sec:code}
\begin{figure}[h]
	\caption{\textbf{CODE inference overview}. When processing a new sample $x$, the confidence 
	measure sums up the average contribution of the detectors from each class weighted by
	the probability of $x$ belonging to that class.}
	\centering
	\includegraphics[width=\textwidth]{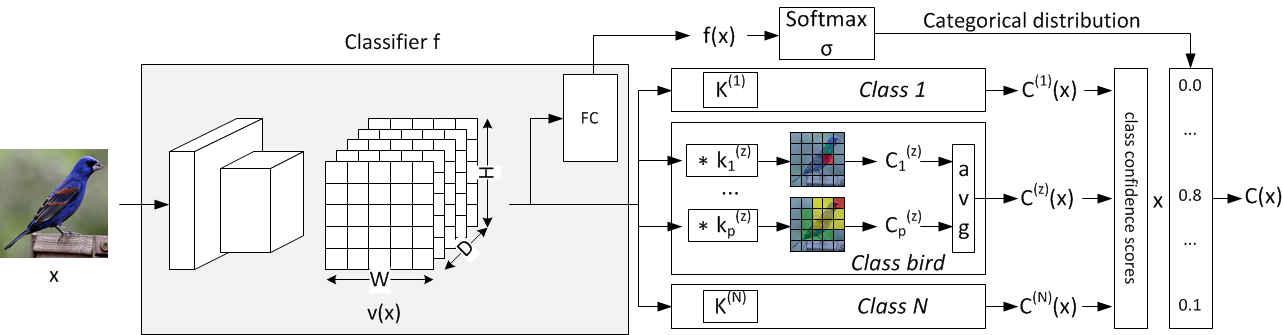}
	\label{fig:CODE_inference}
	\vspace{-0.5cm}
\end{figure}

In this section, we present CODE, our proposal for building a \textit{contextualised} OoD detector. CODE is an extension of the \particul algorithm described in
\cite{xudarme2022particul}, which is intended to mine recurring patterns in the latent
representation of a
set of images processed through a CNN.
Patterns are learnt from the last convolutional layer of the classifier $f$
over the training set $X_{train}$, in a plug-in fashion that does not require
the classifier to be retrained.
Let $v$ be the restriction of classifier $f$ up to its last convolutional layer, \ie $f = l\circ v$,  
where $l$ corresponds to the last pooling layer followed by one or several fully connected layers. 
$\forall x\in \mathcal{X}$, $v(x) \in \mathbb{R}^{H\times W \times D}$ is a convolutional map of
$D$-dimensional vectors. The purpose of the \particul algorithm is to learn $p$
distinct $1\times 1\times D$ convolutional kernels
$K=[k_1, \ldots, k_p]$ (or \emph{detectors}),
such that $\forall x \in X_{train}$: 
1) each kernel $k_i$ strongly correlates with exactly one vector in $v(x)$
(\textit{Locality} constraint); 
2) each vector in $v(x)$ strongly correlates with at most one kernel $k_i$ 
(\textit{Unicity} constraint).
\paragraph{Learning \textit{class-conditional} pattern detectors.} While \particul is an
unsupervised approach restricted to fine-grained recognition datasets, CODE uses the 
training labels from $X_{train}$ to learn $p$ detectors \textit{per class}.
More precisely, let $K^{(c)} = [k_1^{(c)},\ldots k_p^{(c)}]$ be the set of kernel detectors 
for class $c$.
Similar to \cite{xudarme2022particul}, we define the normalised activation map between 
kernel $k_i^{(c)}$ and image $x$ as:
\begin{equation}
	P_i^{(c)}(x) = \sigma\big(v(x)*k_i^{(c)}\big)\in \mathbb{R}^{H\times W}
\end{equation}
where $\sigma$ is the \textit{softmax} normalisation function. 
We also define the cumulative activation map, which sums the normalised scores
for each vector in $v(x)$, \ie
\begin{equation}
	S^{(c)}(x) = \sum\limits_{i=1}^p P_i^{(c)}(x)\in \mathbb{R}^{H\times W}
\end{equation}
Then, we define the \textit{Locality} and \textit{Unicity} objective functions as follows:
\begin{equation}\label{eq:locality}
\mathcal{L}_{l} = -\sum\limits_{(x,y)\in X_{train}} \sum\limits_{c=1}^N\sum\limits_{i=1}^p  \mathds{1}_{[c=y]} \times \max \big(P_i^{(c)}(x)*u\big)
\end{equation}
\begin{equation}\label{eq:unicity}
\mathcal{L}_{c} = \sum\limits_{(x,y)\in X_{train}}\sum\limits_{c=1}^N  \mathds{1}_{[c=y]} \times \max \Big(0, \max \big(S^{(c)}(x)\big)-t\Big)
\end{equation}
where $\mathds{1}$ is the indicator function, and $u$ is a $3\times 3$ uniform kernel that 
serves as a relaxation of the Locality constraint. 
Due to the softmax normalisation of the activation map $P_i^{(c)}(x)$,
$\mathcal{L}_l$ is minimised when, for all images $x$ of class $c$, each kernel $k_i^{(c)}$
strongly correlates with one and only one $3\times 3$ region of the convolutional map $v(x)$.
Meanwhile, $\mathcal{L}_{c}$ is minimised when, for all images $x$ of class $c$, the sum of
normalised correlation scores between a given vector in $v(x)$ and all kernels $k_i^{(c)}$ 
does not exceed a threshold $t=1$, ensuring that no vector in $v(x)$ correlates too strongly 
with multiples kernels. 
The final training objective is $\mathcal{L} = \mathcal{L}_l + \lambda_u\mathcal{L}_u$. 
Importantly, \emph{we do not explicitly specify a target pattern for each detector}, but our
training objective will ensure that we obtain detectors for $p$ different patterns for each
class. Moreover, since patterns may be similar across different classes (e.g., the wheels 
on a car or a bus), we \emph{do not treat images from other classes as negative samples} 
during training.
\paragraph{Confidence measure.} After training, we build our confidence measure using the function 
${H_i^{(c)}(x) = \max\limits_{v^* \in v(x)} (v^**k_i^{(c)})}$
returning the maximum correlation score between kernel
$k_i^{(c)}$ and $v(x)$. Assuming
that each detector will correlate more strongly with images from $\mathcal{D}_{id}$ than
images outside of $\mathcal{D}_{id}$, we first estimate over $X_{train}$ the mean value $\mu_i^{(c)}$ and standard deviation $\sigma_i^{(c)}$ of the distribution of maximum correlation scores $H_i^{(c)}$ for $(x, c) \sim \mathcal{P}_{id}$. Then, we define
\begin{equation}\label{eq:CODE_Ci}
\begin{array}{c}
C^{(c)}(x)=\dfrac{1}{p} \sum\limits_{i=1}^p C_i^{(c)}(x),
	\mbox{with }C_i^{(c)}(x)=sig\Big(\big(H_i^{(c)}(x)-\mu_i^{(c)}\big)/\sigma_i^{(c)}\Big)
\end{array}
\end{equation}
as the \textit{class confidence score} for class $c$. Though it could be confirmed using a KS-test on empirical data, the logistic distribution hypothesis used for $H_i^{(c)}$ - rather of the normal distribution used in \particul~- is primarily motivated by the computational effectiveness\footnote{Although good approximations of the normal CDF using sigmoids exist~\cite{eidous2022approximations}} and the normalisation effect of the sigmoid $sig$ that converts a raw correlation score into a value between $0$ and $1$. 
During inference, for $x\in \mathcal{X}$, the confidence measure $C(x)$ is 
obtained by weighting each class confidence score by the probability that $x$ belongs 
to this class:
\begin{equation}\label{eq:code_confidence}
C(x)=\sum\limits_{c=1}^NC^{(c)}(x)\times P(Y=c~|~X=x)
\end{equation}
where the categorical distribution $P(Y~|~X=x)$ is obtained from the vector of normalised 
logits $n = \sigma\big(f(x)\big)$, as shown in Fig.~\ref{fig:CODE_inference}.
Note that it would be possible to build $C(x)$ using 
only the confidence score of the most probable class $c=\argmax\big(f(x)\big)$. However,
using the categorical distribution allows us to mitigate the model (in)accuracy, 
as we will see in Sec.~\ref{sec:experiments}.
\begin{figure}
	\caption{\textbf{Explanations generated by CODE for ID and OoD samples}. For each image, the 
	classification as ID/OoD rely on the presence/absence of class-specific visual cues 
	extracted from the training set.}\label{fig:CODE_explanation}
	\subfloat[Out-of-distribution image.]{
			\centering
			\includegraphics[width=0.48\linewidth]{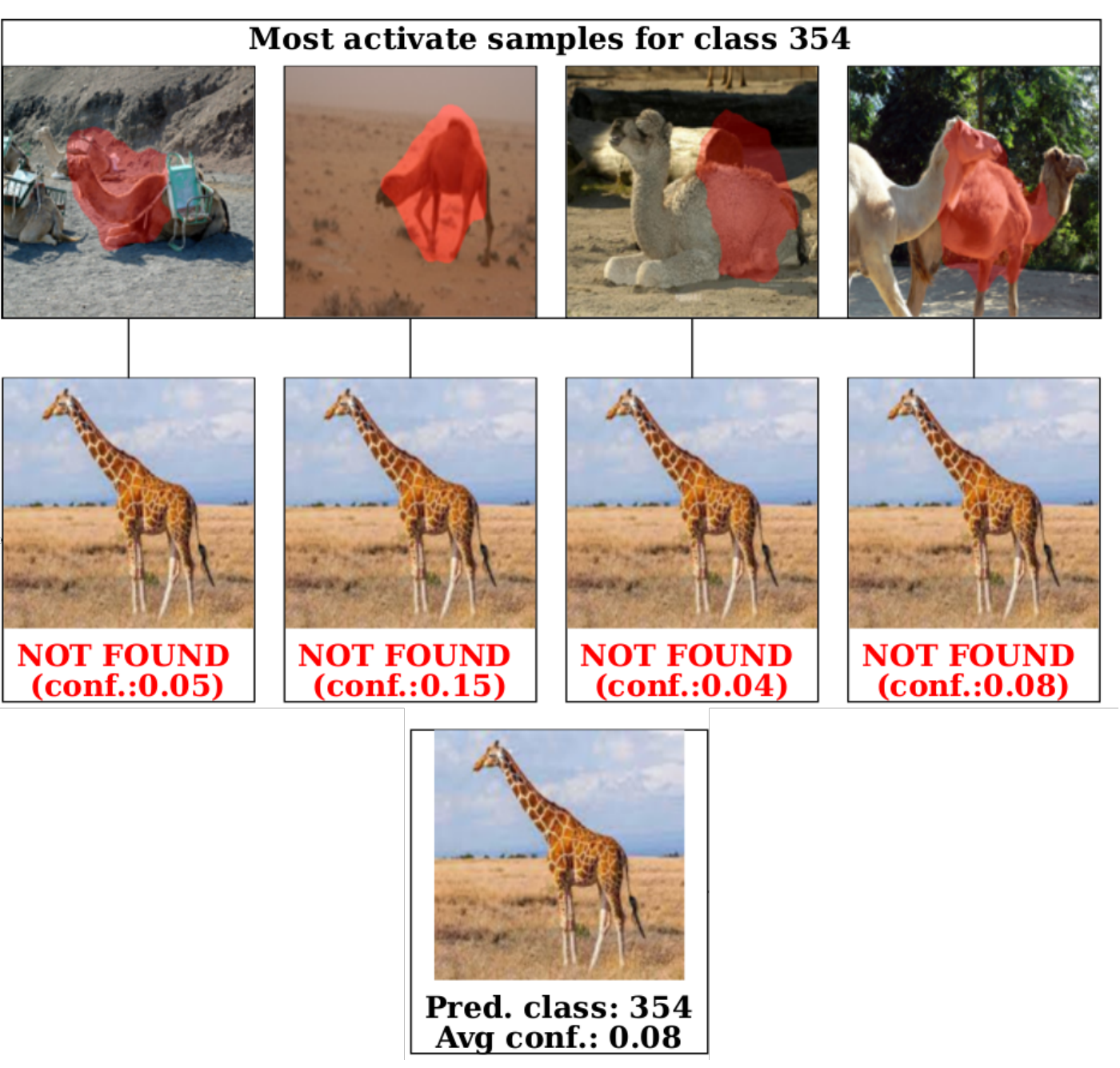}
		
	}
	\hfill
	\subfloat[Inside-Of-Distribution image.]{
			\centering
			\includegraphics[width=0.48\linewidth]{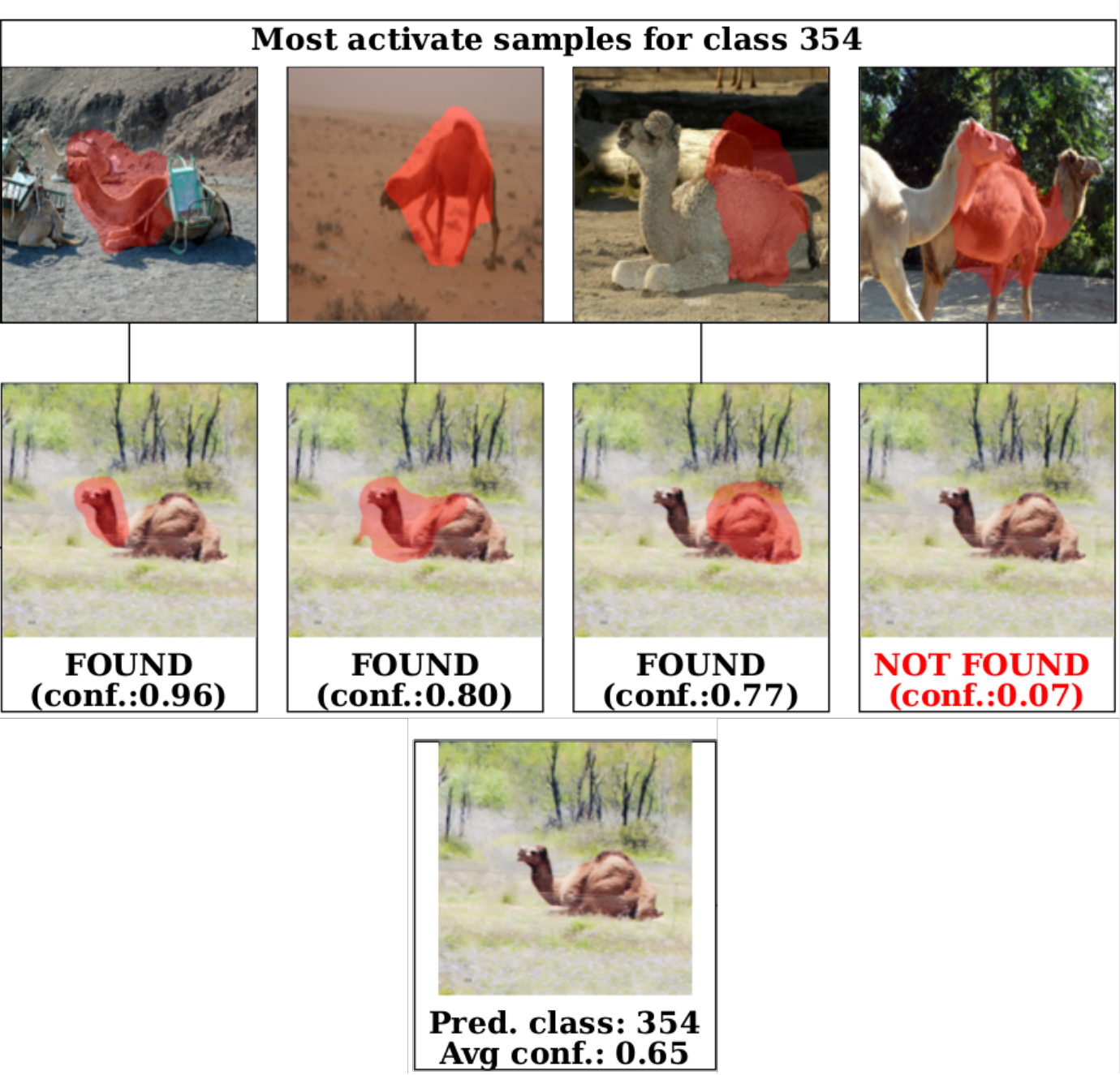}
	}
	\vspace{-0.5cm}
\end{figure}
\paragraph{Extracting examples}
One of the key advantages of CODE over existing OoD detection methods resides in the ability to
provide a visual justification of the confidence measure. For each detection kernel $k_i^{(c)}$ 
for class $c$, we first identify the sample $(x,c)\in X_{train}$ that most faithfully represents
the distribution of correlation scores $H_i^{(c)}$ across the training (in practice, we select
the sample whose correlation score is closest to $\mu_i^{(c)}$). Then, as in \cite{xudarme2022particul}, we locate the pattern associated 
with this detector inside image $x$ using the SmoothGrads~\citep{smilkov2017smoothgrads} algorithm.
This operation
anchors each detector for each class to a part of an image in the training set.
Moreover, the ability to visualise patterns can also serve as a sanity check
to verify that our method has indeed learned unique and \textit{relevant} patterns \wrt to the class object. 

For each new image, as shown in Fig.~\ref{fig:CODE_explanation}, 
we first identify the predicted class ${c=\argmax\big(f(x)\big)}$ returned by the classifier.
Then, we use the individual confidence scores $C_i^{(c)}(x)$ for each detector of class $c$
to infer the presence or absence of each pattern. When the confidence score of a given detector
is above a given threshold (\eg $C_i^{(c)}(x)>0.3$), we highlight the corresponding pattern 
inside image $x$ (again using SmoothGrads) and display the most correlated sample from the training
set as a reference. 
In summary, \emph{we justify the OoD-ness of the new image by pointing out the presence or 
absence of class-specific recurring patterns that were found in the training set}. 
Note that although our confidence measure is computed using \emph{all} class 
confidence scores (weighted by the categorical distribution, see above), we believe that an
explanation built solely on the most probable class can provide enough justification for
the decision, while being sufficiently concise to be understandable by the user.

\section{Experiments}\label{sec:experiments}
In this section, we start by describing the experimental setup designed to answer the following research questions: 1) How does CODE fare against other detection methods on a cross-dataset OoD evaluation benchmark? 2) What is the influence of weighting all class-condition confidence scores (Eq.~\ref{eq:code_confidence}) rather than using only the confidence score of the most probable class? 3) How does the number $p$ of detectors per class influences CODE detection capabilities? 4) How do OoD detection methods behave when applying perturbations on the ID dataset?

\paragraph{Setup}
We performed our evaluation using the OpenOoD framework~\citep{yang2022openood}, which already
implements most recent OoD detection methods. 
For each ID dataset, we used the provided pre-trained classifier for feature
extraction and trained $4$ or $6$ pattern detectors per class, 
using the labels of the training set
and the objective function described in Sec.~\ref{sec:code}. After cross-validation on a CIFAR10 v. CIFAR100 detection benchmark, we set $\lambda_u=1$, putting equal emphasis on the locality and unicity constraints. 
Although CODE trains a high number of 
detectors, the learning process remains computationally efficient since the 
classifier is not modified and only the detectors of the labelled class are updated during the
back-propagation phase. Additionally, for large datasets such as ImageNet, detectors from
different classes can be trained \textit{in parallel} on chunks of the dataset 
corresponding to their respective class. We trained our detectors with RMSprop 
(learning rate $5\times 10^{-4}$, weight decay $10^{-5}$), for 30 epochs (ImageNet) or 200 epochs 
(all other ID datasets). As a comparison, we also implemented a class-based FNRD~\citep{hond2021integrated}, extracting neuron activation values at different
layers of the classifier. 

\paragraph{Cross-dataset OoD evaluation}
\begin{table*}
	\vspace{-0.3cm}
	\caption{\textbf{Comparison of AUROC scores between CODE and state-of-the-art methods on a cross-dataset benchmark.} 
	Results with $^*$ are extracted from~\cite{yang2022openood} - keeping
	only OoD-agnostic methods. We also add
	results of our implementation of a class-based FNRD~\citep{hond2021integrated}.
	Experiments on ImageNet using $6$ CODE detectors have not yet been conducted due to limited ressources (denoted \clock). 
	For readability, AUPR and FPR95 are omitted but available upon request.
	\label{tab:xdataset}\vspace{0.1cm}}
	\centering
{\renewcommand\baselinestretch{1.3}\selectfont
	\resizebox{\textwidth}{!}{
		\begin{tabular}{@{\hskip 8pt}l@{\hskip 6pt}|@{\hskip 6pt}cccc@{\hskip 6pt}|@{\hskip 6pt}c@{\hskip 6pt}|@{\hskip 6pt}cccc@{\hskip 6pt}|@{\hskip 6pt}c@{\hskip 10pt}} 
			\toprule
			& \multicolumn{5}{c@{\hskip 6pt}|@{\hskip 6pt}}{OSR}
			& \multicolumn{5}{c}{OoD Detection (Near-OoD~/~Far-OoD)}
			\\
			& M-6 & C-6 & C-50 & T-20 & Avg.
			& MNIST & CIFAR-10 & CIFAR-100 & ImageNet &  Avg. \\
			\midrule
			MSP$^*$~\citep{hendrycks2017baseline}
			& 96.2 & 85.3 & 81.0 & 73.0 & 83.9
			& 91.5~/~98.5 & 86.9~/~89.6 & 80.1~/~77.6 & 69.3~/~86.2 & 81.9~/~87.9 \\
			ODIN$^*$~\citep{liang2018enhancing}
			& 98.0 & 72.1 & 80.3 & \textbf{75.7} & 81.8
			& 92.4~/~99.0 & 77.5~/~81.9 & 79.8~/~78.5 & 73.2~/~94.4 & 80.7~/~88.4 \\
			MDS$^*$~\citep{lee2018simple}
			& 89.8 & 42.9 & 55.1 & 57.6 & 62.6
			& \textbf{98.0}~/~98.1 & 66.5~/~88.8 & 51.4~/~70.1 & 68.3~/~94.0 & 71.0~/~87.7 \\
			Gram$^*$~\citep{sastry2020detecting}
			& 82.3 & 61.0 & 57.5 & 63.7 & 66.1
			& 73.9~/~\textbf{99.8} & 58.6~/~67.5 & 55.4~/~72.7 & 68.3~/~89.2 & 64.1~/~82.3 \\
			EBO$^*$~\citep{liu2020energy}
			& \textbf{98.1} & 84.9 & 82.7 & 75.6 & 85.3
			& 90.8~/~98.8 & 87.4~/~88.9 & 71.3~/~68.0 & 73.5~/~92.8 & 80.7~/~87.1 \\
			GradNorm$^*$~\citep{huang2021importance}
			& 94.5 & 64.8 & 68.3 & 71.7 & 74.8
			& 76.6~/~96.4 & 54.8~/~53.4 & 70.4~/~67.2 & 75.7~/~95.8 & 69.4~/~78.2 \\
			ReAct$^*$~\citep{sun2021react}
			& 82.9 & 85.9 & 80.5 & 74.6 & 81.0
			& 90.3~/~97.4 & 87.6~/~89.0 & 79.5~/~80.5 & 79.3~/~95.2 & 84.2~/~90.5 \\
			MaxLogit$^*$~\citep{hendrycks2022scaling}
			& 98.0 & 84.8 & 82.7 & 75.5 & 85.3
			& 92.5~/~99.1 & 86.1~/~88.8 & \textbf{81.0}~/~78.6 & 73.6 / 92.3 & 83.3~/~89.7 \\
			KLM$^*$~\citep{hendrycks2022scaling}
			& 85.4 & 73.7 & 77.4 & 69.4 & 76.5
			& 80.3~/~96.1 & 78.9~/~82.7 & 75.5~/~74.7 & 74.2~/~93.1 & 77.2~/~86.7 \\
			ViM$^*$~\citep{wang2022vim}
			& 88.8 & 83.5 & 78.2 & 73.9 & 81.1
    			& 94.6~/~99.0 & 88.0~/~92.7 & 74.9~/~\textbf{82.4} & 79.9~/~\textbf{98.4} & 84.4~/~\textbf{93.1} \\
			KNN$^*$~\citep{sun2022out}
			& 97.5 & \textbf{86.9} &\textbf{83.4} & 74.1 & \textbf{85.5}
			& 96.5~/~96.7 & \textbf{90.5}~/~\textbf{92.8} & 79.9~/~82.2 & \textbf{80.8}~/~98.0 & \textbf{86.9}~/~92.4 \\
			DICE$^*$~\citep{sun2021dice}
			& 66.3 & 79.3 & 82.0 & 74.3 & 75.5
			& 78.2~/~93.9 & 81.1~/~85.2 & 79.6~/~79.0 & 73.8~/~95.7 & 78.2~/~88.3 \\
			FNRD~\citep{hond2021integrated} 
			& 59.4 & 68.2 & 58.4 & 54.3 & 60.1
			& 84.8~/~97.1 & 70.2~/~71.5 & 54.6~/~58.5 & 75.4~/~87.5 & 71.3~/~78.7 \\
			\midrule			
			\multicolumn{9}{l}{\textbf{- This work}} \vspace{.1cm}\\
			\midrule
		 	CODE (p=$4$) & 74.7 & 86.7 & 76.5 & 62.4 & 75.1 &  
		 	 81.8~/~99.5 & 87.8~/~90.7 & 73.9~/~72.4 & 76.6~/~84.4 & 80.0~/~86.8\\
		 	most probable class only  & 73.7 & 86.4 & 74.6 & 61.3 & 74.0 &  
		 	 80.5~/~99.5 & 87.4~/~90.3 & 72.2~/~71.0 & 73.7~/~77.3 & 78.5~/~84.5\\
		 	 \midrule
 		 	CODE (p=$6$) & 73.7 & 86.0 & 76.1 & 61.5 & 74.3 &  
		 	  82.2~/~99.2 & 88.5~/~92.4 & 73.0~/~76.4 & \clock & \\
		 	most probable class only  & 72.8 & 85.7 & 73.9 & 60.4 & 73.2 &  
		 	  81.8~/~98.7 & 87.8~/~91.8 & 70.9~/~74.5 & \clock & \\

			\bottomrule
	\end{tabular}}\par}
	\vspace{-0.6cm}
\end{table*}
The cross-dataset evaluation implemented in OpenOoD
includes a OoD detection benchmark and an Open Set Recognition (OSR) benchmark.
For the OoD detection benchmark, we use the ID/Near-OoD/Far-OoD dataset split proposed
in \cite{yang2022openood}. 

For the OSR benchmark, as in \cite{yang2022openood}, 
M-6 indicates a 6/4 split of MNIST~\citep{deng2012mnist} (dataset split 
between 6 closed set classes used for training and 4 open set classes), C-6 indicates
a 6/4 split of CIFAR10~\citep{krizhevsky09learningmultiple}, C-50 indicates a 50/50 split of CIFAR100~\citep{krizhevsky09learningmultiple} and TIN-20 indicates 
a 20/180 split of TinyImageNet~\citep{krizhevsky2012imagenet}. The AUROC score is averaged over 
5 random splits between closed and open sets. 

\begin{figure}
	\captionof{figure}{\textbf{Evolution of the average confidence score v. magnitude 
	of the perturbation.} \label{fig:perturbation_curve}
	Curves in \textcolor{red}{red} indicate anomalous behaviours. 
	Since all methods have different calibration values, we omit the units on the 
	y-axis, focusing on the general evolution of the average confidence score over the 
	perturbed dataset.}
	\centering
	\includegraphics[width=\linewidth]{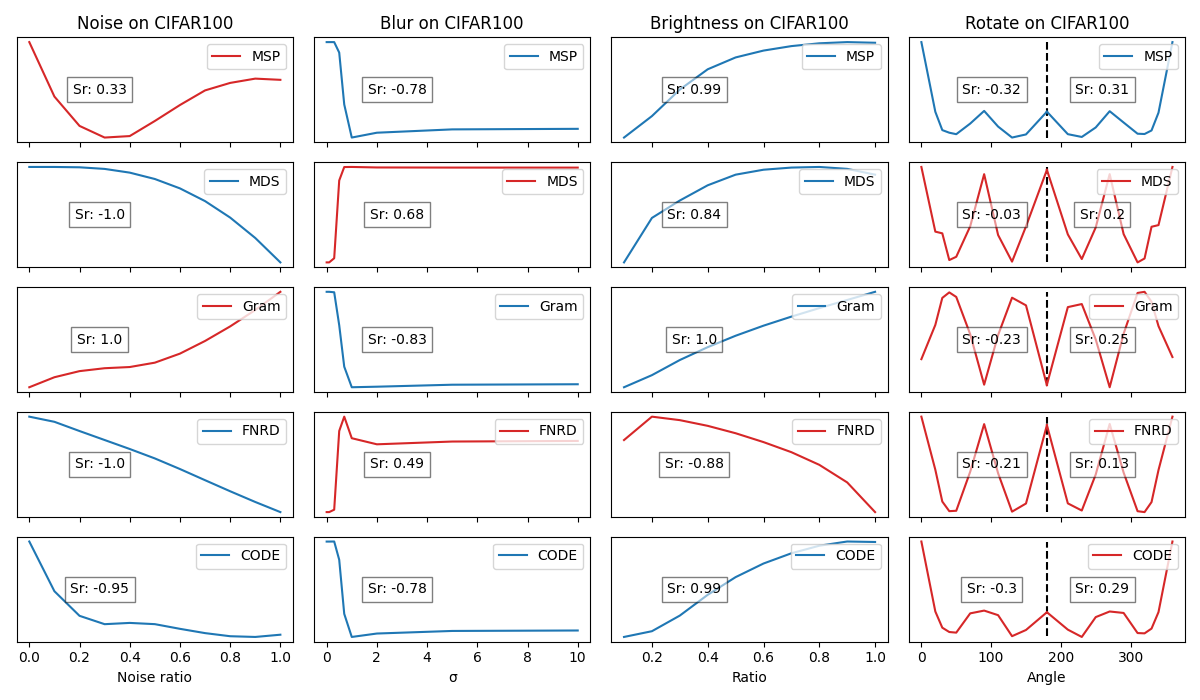}
\end{figure}

The results, summarised in Table~\ref{tab:xdataset}, show that CODE displays OoD detection
capabilities on par with most state-of-the-art methods (top-10 on OSR benchmark, top-8 on Near-OoD detection, top-9 on Far-OoD detection).
Moreover, as discussed in Sec.~\ref{sec:code}, 
using the categorical 
distribution of the output of the classifier to weight class confidence scores 
systematically yields better results than using only the confidence score of the most 
probable class (up to 7\% on the Far-OoD benchmark for ImageNet). Interestingly, increasing the
number of detectors per class from 4 to 6 does not necessarily improve our results. Indeed,
the Unicity constraint (Eq.~\ref{eq:unicity}) becomes harder to satisfy with a higher 
number of detectors and is ultimately detrimental to the Locality constraint
(Eq.~\ref{eq:locality}). 
This experiment also shows that the choice of Near-OoD/Far-OoD datasets in OpenOoD is not 
necessarily reflected by the average AUROC scores. Indeed, for CIFAR100, most methods exhibit
a higher AUROC for Near-OoD datasets than for Far-OoD datasets. This observation highlights the 
challenges of selecting and sorting OoD datasets according to their relative “distance” to the 
ID dataset, without any explicit formal definition of what this distance should be. 
In this regard, our proposed benchmark using perturbations of the ID dataset aims at
providing a quantifiable distance 
between ID and OoD datasets.
\begin{table}[ht]
	\centering
	\caption{\textbf{Summary of the perturbations}, with definition of $\alpha$ and its range. 
	}\label{tab:perturbation_list}
	\centering
	
	\begin{tabular}{l@{\hskip 6pt}@{\hskip 6pt}l@{\hskip 6pt}@{\hskip 6pt}l}
		\toprule
		\multicolumn{1}{c}{Perturbation $P$} & \multicolumn{1}{c}{Description} &  \multicolumn{1}{c}{Range for $\alpha$} \\
		\midrule
		Blur & Gaussian blur with kernel $3\times 3$ & $\alpha\in [0.0, 10]$ \\
		 & and standard deviation $\sigma=\alpha$ \\
		Noise & Gaussian noise with ratio $\alpha$    & $\alpha\in [0, 1.0]$ \\
		Brightness & Blend black image with & $\alpha\in [0.1, 1.0]$\\
		 & ratio $1-\alpha$ \\
		Rotation forth (R+) & Rotation with degree $\alpha$& $\alpha\in [0, 180]$ \\
		Rotation back (R-) & Rotation with degree $\alpha$ & $\alpha\in [180, 360]$\\
		\bottomrule

	\end{tabular}
\vspace{-0.3cm}
\end{table}
\paragraph{Consistency against perturbations} 
We also evaluated all methods on our 
perturbation benchmark (see Sec.~\ref{sec:perturbation}), measuring the Spearman Rank correlation coefficient (SRCC) between the 
magnitude of the perturbation (see Table~\ref{tab:perturbation_list}) and the average
confidence measured on the perturbed dataset.
The results, shown in Table~\ref{tab:pood},
reveal that, on average, \textit{CODE seem to correlate more strongly to the magnitude of the 
perturbation than all other methods}. Moreover, some OoD methods sometimes display 
unexpected behaviours, depending
on the choice of dataset and perturbation, as shown in Fig.~\ref{fig:perturbation_curve}
. In particular,
MSP tends to increase with the noise ratio, hence the success of adversarial attacks~\citep{hein2019why,szegedy2014intriguing}.
Additionally, by construction, any perturbation reducing the amplitude of neuron activation values
(blur, brightness) has the opposite effect of increasing the FNRD.
Gram also increases with the noise ratio and is highly sensitive
to rotations, although we do not have a satisfactory explanation for this particular behaviour.
We also notice that - contrary to our expectations - the average
confidence does not monotonously decrease when rotating images from 0 to 180°: 
all methods show periodic local maximums of the average confidence that may 
indicate a form of invariance of the network \wrt rotations of specific magnitude
(45° for CIFAR10, 90° for CIFAR100/ImageNet, 180° for MNIST). This effect seems 
amplified for CIFAR100 (see Fig.~\ref{fig:perturbation_curve}). 
Finally, we notice that the top-3 methods for Near-OoD detection
(KNN, ViM and ReAct) also strongly correlate with the magnitude of the perturbation, which opens the door to a more in-depth analysis of the relationship between the two benchmarks.

\begin{table}
	\vspace{-0.5cm}
	\captionof{table}{\textbf{Comparison of OoD methods on our perturbation benchmark.} For each perturbation, $\uparrow$ (resp. $\downarrow$) indicates that the average confidence on the perturbed dataset should increase (resp. decrease) with $\alpha$, \ie that the sign of the SRCC should be positive (resp. negative). Results in \textcolor{red}{red} indicate either a weak correlation (absolute value lower than 0.3) or an unexpected sign of the correlation coefficient, \eg the average Gram confidence score increases with the noise ratio on CIFAR100 ($r_s=1.0$) when it should be decreasing. Results in \textbf{bold} indicate a strong expected correlation (absolute value greater than 0.9). The last column represents the average correlation score, taking into account the expected sign of the correlation (results with $^*$ are partial average values). \clock ~indicates a timeout during the experiments.}
	\label{tab:pood}
	\centering 
	{\renewcommand\baselinestretch{1.3}\selectfont \resizebox{\textwidth}{!}{ 
	\begin{tabular}{@{\hskip 8pt}l@{\hskip 6pt}|@{\hskip 6pt}ccccc@{\hskip 6pt}|@{\hskip 6pt}ccccc@{\hskip 6pt}|@{\hskip 6pt}ccccc@{\hskip 6pt}|c}
	\toprule
	& \multicolumn{5}{c@{\hskip 6pt}|@{\hskip 6pt}}{CIFAR10}
& \multicolumn{5}{c@{\hskip 6pt}|@{\hskip 6pt}}{CIFAR100}
& \multicolumn{5}{c}{ImageNet} & Avg.
\\ 
	  & Noise $\downarrow$ & Blur $\downarrow$ & Bright. $\uparrow$ & R+ $\downarrow$ & R- $\uparrow$ & Noise $\downarrow$ & Blur $\downarrow$ & Bright. $\uparrow$ & R+ $\downarrow$ & R- $\uparrow$ & Noise $\downarrow$ & Blur $\downarrow$ & Bright. $\uparrow$ & R+ $\downarrow$ & R- $\uparrow$ \\ 
	  	MSP & \textcolor{red}{-0.22} & -0.88 & \textbf{0.98} & -0.55 & 0.56 & \textcolor{red}{0.33} & -0.78 & \textbf{0.99} & -0.32 & 0.31 & \textcolor{red}{0.71} & \textbf{-1.0} & \textbf{1.0} & -0.77 & 0.85 & 0.54 \\ 
	ODIN & -0.85 & -0.7 & \textcolor{red}{0.18} & \textcolor{red}{-0.15} & \textcolor{red}{0.13} & \textcolor{red}{-0.15} & -0.77 & 0.75 & \textcolor{red}{-0.22} & \textcolor{red}{0.21} & \textcolor{red}{0.12} & -0.87 & \textcolor{red}{0.2} & -0.81 & 0.81 & 0.45 \\ 
	MDS & \textbf{-1.0} & \textcolor{red}{0.41} & 0.84 & \textcolor{red}{-0.03} & \textcolor{red}{0.19} & \textbf{-1.0} & \textcolor{red}{0.68} & 0.84 & \textcolor{red}{-0.03} & \textcolor{red}{0.2} & \textbf{-1.0} & \textcolor{red}{0.98} & \textcolor{red}{-0.35} & \textcolor{red}{-0.16} & \textcolor{red}{0.11} & 0.20 \\ 
	Gram & \textcolor{red}{1.0} & \textbf{-1.0} & \textbf{1.0} & \textcolor{red}{-0.15} & \textcolor{red}{-0.02} & \textcolor{red}{1.0} & -0.83 & \textbf{1.0} & \textcolor{red}{-0.23} & \textcolor{red}{0.25} & \clock & \clock & \clock & \clock & \clock & 0.24$^*$ \\ 
	EBO & -0.62 & -0.88 & \textbf{0.96} & -0.33 & \textcolor{red}{0.29} & -0.32 & -0.78 & \textbf{0.99} & \textcolor{red}{-0.22} & \textcolor{red}{0.22} & \textcolor{red}{0.63} & \textbf{-0.93} & \textbf{1.0} & -0.78 & 0.75 & 0.56 \\ 
	GradNorm & \textcolor{red}{0.05} & -0.69 & \textcolor{red}{-1.0} & \textcolor{red}{-0.04} & \textcolor{red}{-0.01} & -0.71 & -0.78 & 0.88 & -0.32 & 0.31 & \textcolor{red}{0.75} & \textbf{-0.93} & \textbf{1.0} & -0.47 & 0.41 & 0.32 \\ 
	ReAct & -0.44 & -0.88 & \textbf{0.96} & -0.37 & 0.33 & -0.75 & -0.78 & \textbf{0.99} & \textcolor{red}{-0.22} & \textcolor{red}{0.21} & \textcolor{red}{-0.25} & \textbf{-0.95} & \textbf{1.0} & -0.66 & 0.66 & 0.63 \\ 
	MaxLogit & -0.62 & -0.88 & \textbf{0.96} & -0.33 & 0.33 & \textcolor{red}{0.0} & -0.78 & \textbf{0.99} & \textcolor{red}{-0.22} & \textcolor{red}{0.22} & \textcolor{red}{0.65} & \textbf{-0.93} & \textbf{1.0} & -0.78 & 0.78 & 0.54 \\ 
	KLM & \textcolor{red}{-0.1} & \textbf{-0.93} & \textbf{0.95} & -0.53 & 0.44 & \textcolor{red}{-0.01} & -0.83 & \textbf{0.99} & \textcolor{red}{-0.27} & \textcolor{red}{0.26} & \clock & \clock & \clock & \clock & \clock & 0.53$^*$ \\ 
	ViM & -0.78 & -0.83 & \textbf{0.92} & \textcolor{red}{-0.29} & \textcolor{red}{0.3} & \textbf{-1.0} & -0.88 & \textbf{1.0} & \textcolor{red}{-0.18} & 0.38 & \textbf{-1.0} & \textbf{-1.0} & \textbf{1.0} & -0.43 & 0.43 & 0.69 \\ 
	KNN & -0.36 & -0.88 & \textbf{0.99} & -0.46 & 0.4 & \textcolor{red}{-0.02} & -0.79 & \textbf{1.0} & \textcolor{red}{-0.26} & 0.35 & \textbf{-0.99} & \textbf{-1.0} & \textbf{1.0} & -0.5 & 0.5 & 0.63 \\ 
	DICE & \textbf{-0.97} & -0.88 & \textbf{0.92} & -0.46 & 0.37 & \textbf{-0.99} & -0.78 & \textbf{0.99} & -0.32 & \textcolor{red}{0.22} & \textcolor{red}{0.65} & \textbf{-0.93} & \textbf{1.0} & -0.74 & 0.74 & 0.64 \\ 
	FNRD & \textbf{-1.0} & \textcolor{red}{0.58} & \textcolor{red}{-0.99} & \textcolor{red}{-0.11} & \textcolor{red}{0.08} & \textbf{-1.0} & \textcolor{red}{0.49} & \textcolor{red}{-0.88} & \textcolor{red}{-0.21} & \textcolor{red}{0.13} & \textbf{-1.0} & -0.85 & \textbf{0.99} & -0.35 & 0.35 & 0.21 \\ 
	CODE & -0.69 & -0.88 & \textbf{1.0} & -0.5 & 0.35 & \textbf{-0.95} & -0.78 & \textbf{0.99} & \textcolor{red}{-0.3} & \textcolor{red}{0.29} & -0.85 & \textbf{-0.93} & \textbf{1.0} & -0.85 & 0.83 & \textbf{0.75}\\ 
\bottomrule 
	\end{tabular}}\par}
\end{table}
%
\section{Conclusion \& Future Work}\label{sec:conclusion}
In this paper, we have demonstrated how the detection of
recurring patterns can be exploited to develop CODE, an OoD-agnostic method that also enables a form of visualisation of 
the detected patterns. 
We believe that this unique feature can help the developer verify visually the quality of the OoD detection method and therefore can increase the safety of image classifiers. 
More generally, in the future we wish to study more thoroughly how part visualisation can be leveraged to fix or improve the OoD detection method when necessary. For instance, we noticed some redundant parts during our experiments and believe that such redundancy could be identified automatically, and pruned during the training process to produce a more precise representation of each class.
Additionally, providing a form of justification of the OoD-ness of a sample could also increase the \textit{acceptability} of the method from the end-user point of view, a statement that we wish to confirm by conducting a user study in the future.
Our experiments show that CODE offers consistent results on par with state-of-the-art 
methods in the context of 
two different OoD detection benchmarks, including our new OoD benchmark based 
on perturbations of the reference dataset. 
This new benchmark highlights intriguing behaviours by several state-of-the-art methods (\wrt specific types of perturbation) that should be analysed in details.
Moreover, since these perturbations are equivalent to a \textit{controlled covariate shift}, it would
be interesting to evaluate covariate shift detection methods in the same setting.
Finally, note that CODE could be applied to other part detection algorithms, provided that a
confidence measure could be associated with the detected parts.

\subsubsection{Acknowledgements} 
Experiments presented in this paper were carried out using the Grid'5000 testbed, supported by a scientific interest group hosted by Inria and including CNRS, RENATER and several Universities as well as other organisations (see https://www.grid5000.fr).
This work has been partially supported by MIAI@Grenoble Alpes, (ANR-19-P3IA-0003) and TAILOR, a project funded by EU Horizon 2020 research and innovation programme under GA No 952215.

%
%
%

\end{document}